%% file: main.tex
\def\year{2021}\relax
\documentclass[letterpaper]{article} 
\usepackage{aaai21}  
\usepackage{times}  
\usepackage{helvet} 
\usepackage{courier}  
\usepackage[hyphens]{url}  
\usepackage{graphicx} 
\usepackage{natbib}  
\usepackage{caption} 
\usepackage{booktabs}
\usepackage{footnote}

\usepackage{amsmath,amssymb,mathrsfs}
\usepackage[ruled,linesnumbered]{algorithm2e}

\usepackage{epstopdf}
\usepackage{multirow}
\usepackage[skip=0pt]{subcaption}
\usepackage{soul}

\usepackage{tabularx}
 
\usepackage{enumitem}

\usepackage{microtype}
\usepackage[switch]{lineno}

\title{A Simple and Effective Self-Supervised Contrastive Learning Framework\\ for Aspect Detection}
\author{
    Tian Shi\textsuperscript{\rm 1}, Liuqing Li\textsuperscript{\rm 2}, Ping Wang\textsuperscript{\rm 1}, Chandan K. Reddy\textsuperscript{\rm 1}\\
}
\affiliations{
    \textsuperscript{\rm 1}Department of Computer Science, Virginia Tech\\
    \textsuperscript{\rm 2}Verizon Media\\
    tshi@vt.edu, liuqing.li@verizonmedia.com, ping@vt.edu, 
    reddy@cs.vt.edu
}

\begin{document}

\maketitle

\begin{abstract}
Unsupervised aspect detection (UAD) aims at automatically extracting interpretable aspects and identifying aspect-specific segments (such as sentences) from online reviews.
However, recent deep learning based topic models, specifically aspect-based autoencoder, suffer from several problems such as extracting noisy aspects and poorly mapping aspects discovered by models to the aspects of interest. To tackle these challenges, in this paper, we first propose a self-supervised contrastive learning framework and an attention-based model equipped with a novel smooth self-attention (SSA) module for the UAD task in order to learn better representations for aspects and review segments. Secondly, we introduce a high-resolution selective mapping (HRSMap) method to efficiently assign aspects discovered by the model to the aspects of interest. We also propose using a knowledge distillation technique to further improve the aspect detection performance. Our methods outperform several recent unsupervised and weakly supervised approaches on publicly available benchmark user review datasets. Aspect interpretation results show that extracted aspects are meaningful, have a good coverage, and can be easily mapped to aspects of interest. Ablation studies and attention weight visualization also demonstrate effectiveness of SSA and the knowledge distillation method.
\end{abstract}

\input{introduction}

\input{related_work}
\input{model}

\input{experiments}

\input{conclusion}

\section{Acknowledgments}
This work was supported in part by the US National Science Foundation grants  IIS-1707498, IIS-1838730, and NVIDIA Corporation.

\bibliographystyle{aaai}
\bibliography{ref}

\clearpage
\appendix
\input{appendix}

\end{document}

%% file: introduction.tex
\section{Introduction}
\label{sec:intro}

Aspect detection, which is a vital component of aspect-based sentiment analysis \cite{pontiki-etal-2014-semeval,pontiki-etal-2015-semeval}, aims at identifying predefined aspect categories (e.g., \textit{Price}, \textit{Quality}) discussed in segments (e.g., sentences) of online reviews.
Table~\ref{tab:example} shows an example review about a television from several different aspects, such as \textit{Image}, \textit{Sound}, and \textit{Ease of Use}.
With a large number of reviews, automatic aspect detection allows people to efficiently retrieve review segments of aspects they are interested in.
It also benefits many downstream tasks, such as review summarization \cite{angelidis-lapata-2018-summarizing} and recommendation justification \cite{ni-etal-2019-justifying}.

\begin{table}[!t]
    \centering
    \resizebox{\linewidth}{!}{
    \begin{tabular}{m{19em}c}
    \toprule
    \textbf{Sentence} & \textbf{Aspect} \\\hline
    Replaced my 27" jvc clunker with this one. & General \\
    It fits perfectly inside our armoire. & General \\
    Good picture. & Image \\
    Easy to set up and program. & Ease of Use \\
    Descent sound, not great... & Sound \\
    We have the 42" version of this set downstairs. & General \\
    Also a solid set. & General \\
    \bottomrule
    \end{tabular}}
    \caption{An example from Amazon product reviews about a television and aspect annotations for every sentence.}
    \vspace{-4mm}
    \label{tab:example}
\end{table}

There are several research directions for aspect detection.
\textit{Supervised approaches} \cite{zhang2018deep} can leverage annotated labels of aspect categories but suffer from domain adaptation problems \cite{rietzler2020adapt}.
Another research direction consists of \textit{unsupervised approaches} and has gained a lot of attention in recent years.
Early unsupervised systems are dominated by Latent Dirichlet Allocation (LDA) based topic models \cite{brody2010unsupervised,mukherjee2012aspect,GARCIAPABLOS2018127,rakesh2018sparse,zhang2019discovering}.
However, several recent studies have revealed that LDA-based approaches do not perform well for aspect detection and the extracted aspects are of poor quality (incoherent and noisy) \cite{he-etal-2017-unsupervised}.
Compared to LDA-based approaches, 
deep learning models, such as aspect-based autoencoder (ABAE) \cite{he-etal-2017-unsupervised,luo2019unsupervised}, have shown excellent performance in extracting coherent aspects and identifying aspect categories for review segments.
However, these models require some human effort to manually map model discovered aspects to aspects of interest, which may lead to inaccuracies in mapping especially when model discovered aspects are noisy.
Another research direction is based on \textit{weakly supervised approaches} that leverage a small number of aspect representative words (namely, \textit{seed words}) for the fine-grained aspect detection \cite{angelidis-lapata-2018-summarizing,karamanolakis-etal-2019-leveraging}.
Although these models outperform unsupervised approaches, they do make use of human annotated data to extract high-quality aspect seed words, which may limit their application.
In addition, they are not able to automatically discover new aspects from review corpus.

We focus on the problem of unsupervised aspect detection (UAD) since massive amount of reviews are generated every day and many of them are for newer products. It is difficult for humans to efficiently capture new aspects and manually annotate segments for them at scale. Motivated by ABAE, we learn interpretable aspects by mapping aspect embeddings into word embedding space, so that aspects can be interpreted by the nearest words. To learn better representations for both aspects and review segments, we formulate UAD as a self-supervised 
representation learning problem and solve it using a contrastive learning algorithm, which is inspired by the 
success of self-supervised contrastive learning in visual representations \cite{chen2020simple,he2020momentum}.
In addition to the learning algorithm, we also resolve two problems that deteriorate the performance of ABAE, including its self-attention mechanism for segment representations and aspect mapping strategy (i.e., many-to-one mapping from aspects discovered by the model to aspects of interest). Finally, we discover that the quality of aspect detection can be further improved by knowledge distillation \cite{hinton2015distilling}.
The contributions of this paper are summarized as follows:
\begin{itemize}[leftmargin=*,topsep=0pt,itemsep=1pt,partopsep=1pt, parsep=1pt]
    \item Propose a self-supervised contrastive learning framework for the unsupervised aspect detection task.
    \item Introduce a high-resolution selective mapping strategy to map model discovered aspects to the aspects of interest.
    \item Utilize knowledge distillation to further improve the performance of aspect detection.
    \item Conduct systematic experiments on seven benchmark datasets and demonstrate the effectiveness of our models both quantitatively and qualitatively.
 \end{itemize}


%% file: related_work.tex
\section{Related Work}
\label{sec:related}

Aspect detection is an important problem of aspect-based sentiment analysis \cite{zhang2018deep,shi2019document}.
Existing studies attempt to solve this problem in several different ways, including rule-based, supervised, unsupervised, and weakly supervised approaches.
\textit{Rule-based approaches} focus on lexicons and dependency relations, and utilize manually defined rules to identify patterns and extract aspects \cite{qiu2011opinion,liu2016improving}, which require domain-specific knowledge or human expertise.
\textit{Supervised approaches} usually formulate aspect extraction as a sequence labeling problem that can be solved by hidden Markov models (HMM) \cite{jin2009opinionminer}, conditional random fields (CRF) \cite{li2010structure,mitchell2013open,yang2012extracting}, and recurrent neural networks (RNN) \cite{wang2016recursive,liu2015fine}.
These approaches have shown better performance compared to the rule-based ones, but require large amounts of labeled data for training. 
\textit{Unsupervised approaches} do not need labeled data.
Early unsupervised systems are dominated by Latent Dirichlet Allocation (LDA)-based topic models \cite{brody2010unsupervised,zhao2010jointly,chen2014aspect,GARCIAPABLOS2018127,shi2018short}.
\citeauthor{wang-etal-2015-sentiment} \shortcite{wang-etal-2015-sentiment} proposed a restricted Boltzmann machine (RBM) model to jointly extract aspects and sentiments.
Recently, deep learning based topic models \cite{akash2017autoencoding,luo2019unsupervised,he-etal-2017-unsupervised} have shown strong performance in extracting coherent aspects.
Specifically, aspect-based autoencoder (ABAE) \cite{he-etal-2017-unsupervised} and its variants
\cite{luo2019unsupervised} have also achieved competitive results in detecting aspect-specific segments from reviews.
The main challenge is that they need some human effort for aspect mapping.
\citeauthor{tulkens-van-cranenburgh-2020-embarrassingly} \shortcite{tulkens-van-cranenburgh-2020-embarrassingly} propose a simple heuristic model that can use nouns in the segment to identify and map aspects, however, it strongly depends on the quality of word embeddings, and its applications have so far been limited to restaurant reviews.
\textit{Weakly-supervised approaches} usually leverage aspect seed words as guidance for aspect detection \cite{angelidis-lapata-2018-summarizing,karamanolakis-etal-2019-leveraging,zhuang2020joint} and achieve better performance than unsupervised approaches.
However, most of them rely on human annotated data to extract high-quality seed words and are not flexible to discover new aspects from a new corpus.
In this paper, we are interested in unsupervised approaches for aspect detection and dedicated to tackle challenges in aspect learning and mapping.

%% file: model.tex
\section{The Proposed Framework}
\label{sec:models}

\begin{figure}[!tp]
	\centering
	\includegraphics[width=0.47\textwidth]{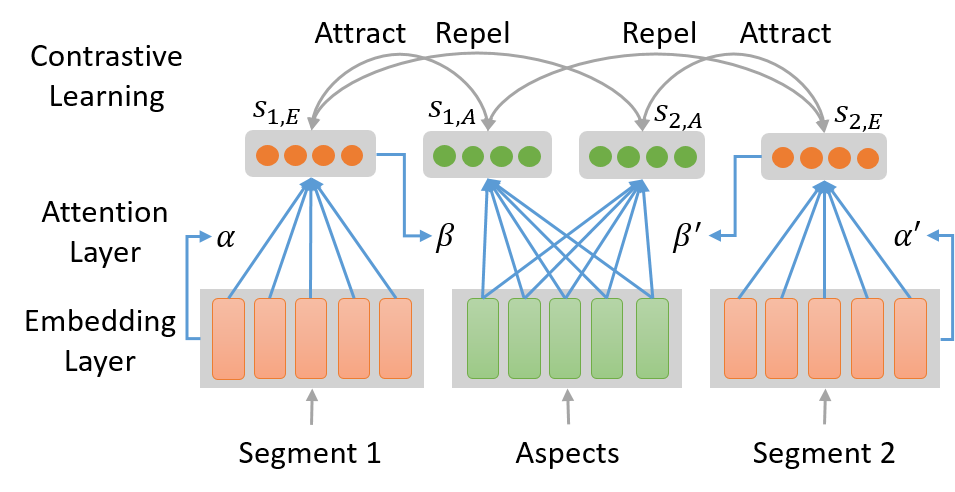}
	\caption{The proposed self-supervised contrastive learning framework. Attract and Repel represent positive and negative pairs, respectively.}
	\vspace{-3mm}
	\label{fig:model_struct}
\end{figure}

In this section, we describe our self-supervised contrastive learning framework for aspect detection shown in Fig.~\ref{fig:model_struct}.
The goal is to first learn a set of interpretable aspects (named as \textit{model-inferred aspects}), and then extract aspect-specific segments from reviews so that they can be used in downstream tasks.

\subsubsection{Problem Statement} The \textit{Aspect detection problem} is defined as follows: 
given a review segment $x=\{x_1, x_2,...,x_T\}$ such as a sentence or an elementary discourse unit (EDU)~\cite{mann1988rhetorical}, the goal is to predict an aspect category $y_k\in \{y_1,y_2,...,y_K\}$, where $x_t$ is the index of a word in the vocabulary, $T$ is the total length of the segment, $y_k$ is an aspect among all aspects that are of interest (named as \textit{gold-standard aspects}), and $K$ is the total number of gold-standard aspects.
For instance, when reviewing restaurants, we may be interested in the following gold-standard aspects: \textit{Food}, \textit{Service}, \textit{Ambience}, etc. Given a review segment, it most likely relates to one of the above aspects.

The first challenge in this problem is to learn model-inferred aspects from unlabeled review segments and map them to a set of gold-standard aspects.
Another challenge is to accurately assign each segment in a review to an appropriate gold-standard aspect $y_k$.
For example, in restaurants reviews, ``\textit{The food is very good, but not outstanding.}"$\rightarrow$\textit{Food}.
Therefore, we propose a series of modules in our framework, including segment representations, contrastive learning, aspect interpretation and mapping, and knowledge distillation, to overcome both challenges and achieve our goal.

\subsection{Self-Supervised Contrastive Learning (SSCL)}
To automatically extract interpretable aspects from a review corpus, a widely used strategy is to learn aspect embeddings in the word embedding space so that the aspects can be interpreted using their nearest words \cite{he-etal-2017-unsupervised,angelidis-lapata-2018-summarizing}.
Here, we formulate this learning process as a \textit{self-supervised representation learning} problem.

\subsubsection{Segment Representations}
For every review segment in a corpus, we construct two representations directly based on (i) word embeddings and (ii) aspect embeddings. 
Then, we develop a \textit{contrastive learning mechanism} to map aspect embeddings to the word embedding space.
Let us denote a word embedding matrix as $E\in\mathbb{R}^{V\times M}$, where 
$V$ is the vocabulary size and $M$ is the dimension of word vectors.
The aspect embedding matrix is represented by $A\in\mathbb{R}^{N\times M}$, where $N$ is the number of model-inferred aspects.

Given a review segment $x=\{x_1,x_2,...,x_T\}$, we construct a vector representation $s_{x,E}$ based on its word embeddings $\{E_{x_1},E_{x_2},..., E_{x_T}\}$, along with a novel self-attention mechanism, i.e.,
\begin{equation}
    s_{x,E}=\sum_{t=1}^T\alpha_t E_{x_t},\\[0.2em]
    \label{eqn:sxe}
\end{equation}
where $\alpha_t$ is an attention weight and is calculated as follows:
\begin{equation}
\aligned
&\alpha_t=\frac{\exp{(u_t)}}{\sum_{\tau=1}^T\exp{(u_\tau)}}\\[0.5em] &u_t=\lambda\cdot\tanh{(q^\top\left(W_EE_{x_t}+b_E)\right)}
\endaligned
\label{eqn:smoothattn}
\end{equation}
Here, $u_t$ is an alignment score and  $q=\frac{1}{T}\sum_{t=1}^TE_{x_t}$ is a query vector. $W_E\in\mathbb{R}^{M\times M}$, $b_E\in\mathbb{R}^{M}$ are trainable parameters, and the smooth factor $\lambda$ is a hyperparameter.
More specifically, we call this attention mechanism as \textbf{Smooth Self-Attention (SSA)}. 
It applies an activation function $\tanh$ to prevent the model from using a single word to represent the segment, thus increasing the robustness of our model.
For example, for the segment ``\textit{plenty of ports and settings}'', SSA will attend both ``\textit{ports}'' and ``\textit{settings}'', while regular self-attention may only concentrate on ``\textit{settings}''.
Hereafter, we will use \textbf{RSA} to represent regular self-attention adopted in \cite{angelidis-lapata-2018-summarizing}.
In our experiments, we discover that RSA without smoothness gets worse performance compared to a simple average pooling mechanism.

Further, we also construct a vector representation $s_{x,A}$ for the segment $x$ with global aspect embeddings $\{A_1, A_2,...,A_N\}$ through another attention mechanism, i.e.,
\begin{equation}
    s_{x,A}=\sum_{n=1}^N\beta_n A_n
    \label{eqn:sxa} \\[0.3em]
\end{equation}
The attention weight $\beta_n$ is obtained by
\begin{equation}
    \beta_n=\frac{\exp{(v_{n,A}^\top s_{x,E}+b_{n,A})}}{\sum_{\eta=1}^N\exp{(v_{\eta,A}^\top s_{x,E}+b_{\eta,A})}},  \\[0.3em]
\end{equation}
where $v_{n,A}\in\mathbb{R}^M$ and $b_{n,A}\in\mathbb{R}$ are learnable parameters.
$\beta=\{\beta_1,\beta_2,...,\beta_N\}$ can be also interpreted as \textbf{soft-labels (probability distribution) over model-inferred aspects} for a review segment.

\IncMargin{1.2em}
\begin{algorithm}[tp]
	\SetAlgoLined
	\SetKw{Initialize}{Initialize}
	\SetKw{Define}{Define}
	\KwIn{Batch size $X$; constants $\lambda$ and $\tau$; network structures\;}
	\KwOut{Aspect embedding matrix $A$; model parameters $W_E$, $b_E$, $v_{A}$, $b_{A}$\;}
	\BlankLine
	\Initialize{Matrix $E$ with pre-trained word vectors; matrix $A$ with k-means centroids\;}
	\For{sampled mini-batch of size $X$}{
	    \For{i=1,X}{
		    Calculate $s_{i,E}$ with Eq.~(\ref{eqn:sxe})\;
		    Calculate $s_{i,A}$ with Eq.~(\ref{eqn:sxa})\;
		}
		\For{i=1,X; j=1,X}{
		    Calculate $\text{sim}(s_{j,E},s_{i,A})$ with Eq.~(\ref{eqn:sim})\;
		}
		\For{i=1,X}{
		    Calculate $l_i$ with Eq.~(\ref{eqn:loss_single})\;
		}
		Calculate regularization term $\Omega$ using Eq.~(\ref{eqn:regularization})\;
        \Define{Loss function $\mathcal{L}=\frac{1}{X}\sum_{i=1}^X l_i$ + $\Omega$\;}
        Update learnable parameters to minimize $\mathcal{L}$.
	}
	\caption{The SSCL Algorithm}
	\label{alg:sscl}
\end{algorithm}
\DecMargin{1.2em}

\subsubsection{Contrastive Learning}

Inspired by recent contrastive learning algorithms \cite{chen2020simple}, SSCL learns aspect embeddings by introducing a contrastive loss to maximize the agreement between two representations of the same review segment.
During training, we randomly sample a mini-batch of $X$ examples and define the contrastive prediction task on pairs of segment representations from the mini-batch, which is denoted by $\{(s_{1,E},s_{1,A}),(s_{2,E},s_{2,A}),...(s_{X,E},s_{X,A})\}$.
Similar to \cite{chen2017sampling}, we treat $(s_{i,E},s_{i,A})$ as a positive pair and  $\{(s_{j,E},s_{i,A})\}_{j\neq i}$ as negative pairs within the mini-batch.
The contrastive loss function for a positive pair of examples is defined as
\begin{equation}
    l_i=-\log\frac{\exp{(\text{sim}(s_{i,E},s_{i,A})}/\mu)}{\sum_{j=1}^X\mathbb{I}_{[j\neq i]}\exp{(\text{sim}(s_{j,E},s_{i,A})}/\mu)},
    \label{eqn:loss_single}
\end{equation}
where $\mathbb{I}_{[j\neq i]}\in\{0,1\}$ is an indicator function that equals $1$ iff $j\neq i$
and $\mu$ represents a temperature hyperparameter. We utilize cosine similarity to measure the similarity between $s_{j,E}$ and $s_{i,A}$, which is calculated as follows:
\begin{equation}
    \text{sim}(s_{j,E},s_{i,A})=\frac{(s_{j,E})^\top s_{i,A}}{\|s_{j,E}\|\|s_{i,A}\|},
    \label{eqn:sim}
\end{equation}
where $\|\cdot\|$ denotes $L_2$-norm.

We summarize our SSCL framework in Algorithm~\ref{alg:sscl}.
Specifically, in line 1, the aspect embedding matrix $A$ is initialized with the centroids of clusters by running k-means on the word embeddings.
We follow \cite{he-etal-2017-unsupervised} to penalize the aspect embedding matrix and ensure diversity of different aspects.
In line 13, the regularization term $\Omega$ is defined as
\begin{equation}
    \Omega=\|\mathcal{A}\mathcal{A}^\top-I\|,
\label{eqn:regularization}
\end{equation}
where each row of matrix $\mathcal{A}$, denoted by $\mathcal{A}_j$, is obtained by normalizing the corresponding row in $A$, i.e., $\mathcal{A}_j=A_j/\|A_j\|$.

\subsection{Aspect Interpretation and Mapping}

\subsubsection{Aspect Interpretation}
\label{sec:asp_interpretation}
In the training stage, we map aspect embeddings to the word embedding space in order to extract interpretable aspects.
With embedding matrices $A$ and $E$, we first calculate a similarity matrix 
$$
G=AE^\top,
$$ 
where $G\in\mathbb{R}^{N\times V}$.
Then, we use the top-ranked words based on $G_n$ to represent and interpret each model-inferred aspect $n$.
In our experiments, the matrix with inner product similarity produces more meaningful representative words compared to using the cosine similarity (see Table~\ref{tab:aspect-words}).

\begin{figure}[!t]
	\centering
	\includegraphics[width=0.49\textwidth]{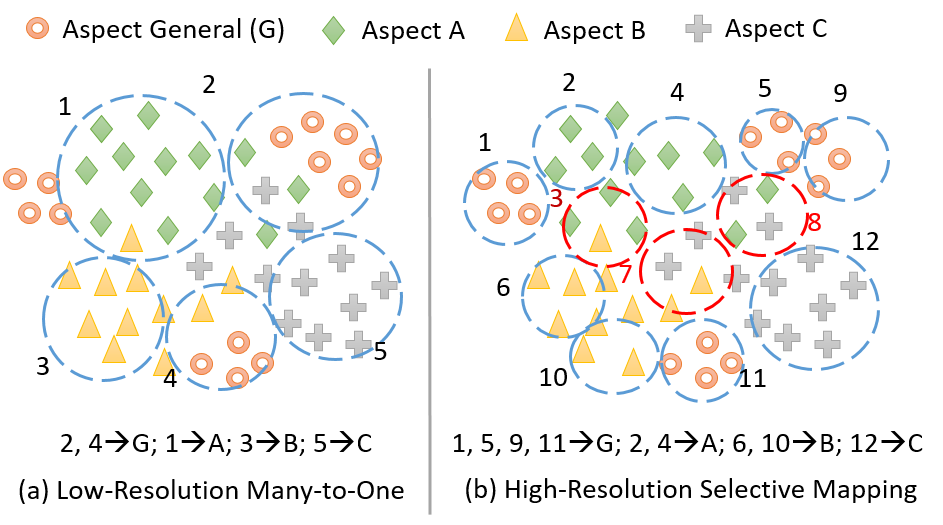}
	\caption{Comparison of aspect mappings. For HRSMap, aspects 3, 7, and 8 are not mapped to gold-standard aspects.}
	\vspace{-3mm}
	\label{fig:aspect_mapping}
\end{figure}

\subsubsection{Aspect Mapping} 

Most unsupervised aspect detection methods focus on the coherence and meaningfulness of model-inferred aspects, and prefer to map every model-inferred aspect (\textbf{MIA}) to a gold-standard aspect (\textbf{GSA}) \cite{he-etal-2017-unsupervised}.
Here, we call this mapping as \textbf{many-to-one mapping}, since the number of model-inferred aspects are usually larger than the number of gold-standard aspects.
Weakly supervised approaches leverage human-annotated datasets to extract the aspect representative words, so that model-inferred aspects and gold-standard aspects have \textbf{one-to-one mapping} \cite{angelidis-lapata-2018-summarizing}.
Different from the two mapping strategies described above, we propose a \textbf{high-resolution selective mapping (HRSMap)} strategy as shown in Fig.~\ref{fig:aspect_mapping}. Here, high-resolution means that the number of model-inferred aspects should be at least 3 times more than the number of gold-standard aspects, so that model-inferred aspects have a better coverage. Selective mapping means noisy or meaningless aspects will not be mapped to gold-standard aspects.

In our experiments, we set the number of MIAs to 30, considering the balance between aspect coverage and human-effort to manually map them to GSAs\footnote{Usually, it takes less than 15 minutes to assign 30 MIAs to GSAs.}.
First, we automatically generate keywords of MIAs based aspect interpretation results, where the number of the most relevant keywords for each aspect is set to 10.
Second, we create several rules for aspect mapping: (i) If keywords of a MIA are clearly related to one specific GSA (not \textit{General}), we map this MIA to the GSA.
For example, we map ``\textit{apps, app, netflix, browser, hulu, youtube, stream}'' to \textit{Apps/Interface} (see Table~\ref{tab:aspect-words}).
(ii) If keywords are coherent but not related to any specific GSA, we map this MIA to \textit{General}.
For instance, we map ``\textit{pc, xbox, dvd, ps3, file, game}'' to \textit{General}.
(iii) If keywords are related to more than one GSA, we treat this MIA as a noisy aspect and it will not be mapped.
For example, ``\textit{excellent, amazing, good, great, outstanding, fantastic, impressed, superior}'' may be related to several different GSAs.
(iv) If keywords are not quite meaningful, their corresponding MIA will not be mapped.
For instance, ``\textit{ago, within, last 30, later, took, couple, per, every}'' is a meaningless MIA.
Third, we further verify the quality of aspect mapping using development sets.

Given the soft-labels of model-inferred aspects $\beta$, we calculate soft-labels $\gamma=\{\gamma_1,\gamma_2,...,\gamma_K\}$ over gold-standard aspects for each review segment as follows:
\begin{equation}
    \gamma_k=\sum_{n=1}^N\mathbb{I}_{[f(\beta_n)=\gamma_k]}\beta_n,
\end{equation}
where $f(\beta_n)$ is the aspect mapping for model-inferred aspect $n$.
The hard-label $\hat{y}$ of gold-standard aspects for the segment is obtained by 
\begin{equation}
    \hat{y}=\text{argmax}\{\gamma_1,\gamma_2,...\gamma_K\},
\end{equation}
which can be converted to a one-hot vector with length $K$.

\subsection{Knowledge Distillation}

Given both soft- and hard-labels of gold-standard aspects for review segments, we utilize a simple knowledge distillation method, which can be viewed as \textbf{classification on noisy labeled data}.
We construct a simple classification model, which consists of a segment encoder such as BERT encoder \cite{devlin2019bert}, a smooth self-attention layer (see Eq.~(\ref{eqn:smoothattn})), and a classifier (i.e., a single-layer feed-forward network followed by a softmax activation).
This model is denoted by SSCLS, where the last S represents \textbf{student}.
SSCLS learns knowledge from the \textbf{teacher} model, i.e., SSCL.
The loss function is defined as
\begin{equation}
    \mathcal{L}=-\frac{1}{K}\sum_{k=1}^K \mathbb{I}_{[H(\gamma)<\xi_k]}\cdot\hat{y}_k\log(y_k),
\end{equation}
where $y_k$ is the probability of aspect $k$ predicted by SSCLS. $\hat{y}_k$ is a hard-label given by SSCL.
$H(\gamma)$ represents the Shannon entropy for the soft-labels and is calculated by $H=-\sum_{k=1}^K \gamma_k\log(\gamma_k)$.
Here, the scalar $\xi_k=\chi_\text{G}$ if aspect $k$ is \textit{General} and $\xi_k=\chi_\text{NG}$, otherwise. 
Both $\chi_\text{G}$ and $\chi_\text{NG}$ are hyperparameters.
Hereafter, we will refer to $\mathbb{I}_{[H(\gamma)<\xi_k]}$ as an \textbf{Entropy Filter}.

Entropy scores have been used to evaluate the confidence of predictions \cite{mandelbaum2017distance}.
In the training stage, we set thresholds to filter out training samples with low confidence predictions from the SSCL model, thus allowing the student model to focus on training samples for which the model prediction are more confident. Moreover, the student model also benefits from pre-trained encoders and overcomes the disadvantages of data pre-processing for SSCL, since we have removed out-of-vocabulary words and punctuation, and lemmatized tokens in SSCL.
Therefore, SSCLS achieves better performance in segment aspect predictions compared to SSCL.

%% file: experiments.tex
\section{Experiments}
\label{sec:exp}

\subsection{Datasets}

\begin{table}[!t]
    \centering
    \resizebox{\linewidth}{!}{
    \begin{tabular}{c|m{18em}}
    \toprule
    \bf Domains & \bf Aspects \\
    \hline
    Bags & Compartments, Customer Service, Handles, Looks, Price, Quality, Protection, Size/Fit, General.\\
    \hline
    Bluetooth & Battery, Comfort, Connectivity, Durability, Ease of Use, Look, Price, Sound, General\\
    \hline
    Boots & Color, Comfort, Durability, Look, Materials, Price, Size, Weather Resistance, General\\
    \hline
    Keyboards & Build Quality, Connectivity, Extra Function, Feel Comfort, Layout, Looks, Noise, Price, General\\
    \hline
    TVs & Apps/Interface, Connectivity, Customer Service, Ease of Use, Image, Price, Size/Look, Sound, General\\
    \hline
    Vacuums & Accessories, Build Quality, Customer Service, Ease of Use, Noise, Price, Suction Power, Weight, General\\
    \bottomrule
    \end{tabular}}
    \caption{The annotated aspects for Amazon reviews across different domains.}
    \label{tab:anno_aspects}
\end{table}

We train and evaluate our methods on seven datasets: Citysearch restaurant reviews \cite{ganu2009beyond} and Amazon product reviews \cite{angelidis-lapata-2018-summarizing} across six different domains, including Laptop Cases (Bags), Bluetooth Headsets (B/T), Boots, Keyboards (KBs), Televisions (TVs), and Vacuums (VCs).

The Citysearch dataset only has training and testing sets.
To avoid optimizing any models on the testing set, we use restaurant subsets of SemEval 2014 \cite{pontiki-etal-2014-semeval} and SemEval 2015 \cite{pontiki-etal-2015-semeval} datasets as a development set, since they adopt the same aspect labels as Citysearch.
Similar to previous work \cite{he-etal-2017-unsupervised}, we select sentences that only express one aspect, and disregard those with multiple and no aspect labels.
We have also restricted ourselves to three labels (Food, Service, and Ambience), to form a fair comparison with prior work \cite{tulkens-van-cranenburgh-2020-embarrassingly}.
Amazon product reviews are obtained from the OPOSUM dataset \cite{angelidis-lapata-2018-summarizing}. 
Different from Citysearch, EDUs \cite{mann1988rhetorical} are used as segments and each domain has eight representative aspect labels as well as aspect \textit{General} (see Table~\ref{tab:anno_aspects}).

In order to train \textit{SSCL}, all reviews are preprocessed by removing punctuation, stop-words, and less frequent words (\textless10).
For Amazon reviews, reviews are segmented into elementary discourse units (EDUs) through a Rhetorical Structure Theory parser \cite{feng2014linear}.
We have converted EDUs back to sentences to avoid training word2vec \cite{mikolov2013distributed} on very short segments.
However, we still use EDU-segments for training and evaluating different models following previous work \cite{angelidis-lapata-2018-summarizing}.
Table~\ref{tab:statistics} shows statistics of different datasets.

\begin{table}[tp]
    \centering
    \resizebox{\linewidth}{!}{
    \begin{tabular}{crrrrr}
    \toprule
        \bf Dataset & \bf Vocab & \bf W2V & \bf Train & \bf Dev & \bf Test \\\hline
        Citysearch & 9,088 & 279,862 & 279,862 & 2,686 & 1,490 \\
        Bags & 6,438 & 244,546 & 584,332 & 598 & 641 \\
        B/T & 9,619 & 573,206 & 1,419,812 & 661 & 656 \\
        Boots & 6,710 & 408,169 & 957,309 & 548 & 611 \\
        KBs & 6,904 & 241,857 & 603,379 & 675 & 681 \\
        TVs & 10,739 & 579,526 & 1,422,192 & 699 & 748 \\
        VCs & 9,780 & 588,369 & 1,453,651 & 729 & 725 \\
    \bottomrule
    \end{tabular}}
    \caption{The vocabulary size and the number of segments in each dataset. \textbf{Vocab} and \textbf{W2V} represent vocabulary size and word2vec, respectively. Refer to Appendix for more details.}
    \label{tab:statistics}
    \vspace{-3mm}
\end{table}

\begin{table*}[!tp]
    \centering
    \begin{tabular}{lccccccc}
         \toprule
         \bf Methods & \bf Bags & \bf B/T & \bf Boots & \bf KBs & \bf TVs & \bf VCs & \bf AVG \\\hline
         \multicolumn{8}{c}{Unsupervised Methods}\\
         ABAE (\citeyear{he-etal-2017-unsupervised}) & 38.1 & 37.6 & 35.2 & 38.6 & 39.5 & 38.1 & 37.9 \\
         ABAE + HRSMap & 54.9 & 62.2 & 54.7 & 58.9 & 59.9 & 54.1 & 57.5 \\\hline
         \multicolumn{8}{c}{Weakly Supervised Methods}\\
         ABAE$_{init}$ (\citeyear{angelidis-lapata-2018-summarizing}) & 41.6 & 48.5 & 41.2 & 41.3 & 45.7 & 40.6 & 43.2 \\
         MATE (\citeyear{angelidis-lapata-2018-summarizing}) & 46.2 & 52.2 & 45.6 & 43.5 & 48.8 & 42.3 & 46.4 \\
         MATE-MT (\citeyear{angelidis-lapata-2018-summarizing}) & 48.6 & 54.5 & 46.4 & 45.3 & 51.8 & 47.7 & 49.1 \\
         TS-Teacher (\citeyear{karamanolakis-etal-2019-leveraging}) & 55.1 & 50.1 & 44.5 & 52.0 & 56.8 & 54.5 & 52.2\\
         TS-Stu-W2V (\citeyear{karamanolakis-etal-2019-leveraging}) & 59.3 & 66.8 & 48.3 & 57.0 & 64.0 & 57.0 & 58.7 \\
         TS-Stu-BERT (\citeyear{karamanolakis-etal-2019-leveraging}) & 61.4 & 66.5 & 52.0 & 57.5 & 63.0 & 60.4 & 60.2 \\\hline
         SSCL & 61.0 & 65.2 & 57.3 & 60.6 & 64.6 & 57.2 & 61.0\\
         SSCLS-BERT & \bf 65.5 & \bf 69.5 & 60.4 & \bf 62.3 & \bf 67.0 & \bf 61.0 & \bf 64.3\bf \\
         SSCLS-DistilBERT & 64.7 & 68.4 & \bf 61.0 & 62.0 & 66.3 & 59.9 & 63.7\\
         \bottomrule
    \end{tabular}
    \caption{Micro-averaged F1 scores for 9-class EDU-level aspect detection in Amazon reviews. \textbf{AVG} denotes the average of F1 scores across all domains.}
    \label{tab:perform_amazon}
    \vspace{-3mm}
\end{table*}

\begin{table*}[!t]
    \centering
    \begin{tabular}{l|ccc|ccc|ccc|ccc}
        \toprule
        & \multicolumn{3}{c|}{\textbf{Food}}& \multicolumn{3}{c|}{\textbf{Staff}}& \multicolumn{3}{c|}{\textbf{Ambience}}&\multicolumn{3}{c}{\textbf{Overall}}\\\hline
        \bf Methods & \bf P & \bf R & \bf F  & \bf P & \bf R & \bf F & \bf P & \bf R & \bf F & \bf P & \bf R & \bf F \\\hline
        
        SERBM (\citeyear{wang-etal-2015-sentiment}) & 89.1 & 85.4 & 87.2 & 81.9 & 58.2 & 68.0 & 80.5 & 59.2 & 68.2  &  86.0 & 74.6 & 79.5 \\
        ABAE (\citeyear{he-etal-2017-unsupervised}) & 95.3 & 74.1 & 82.8 & 80.2 & 72.8 & 75.7  & 81.5 & 69.8 & 74.0  &  89.4 & 73.0 & 79.6 \\
        W2VLDA (\citeyear{GARCIAPABLOS2018127}) & 96.0 & 69.0 & 81.0 & 61.0 & 86.0 & 71.0 & 55.0 & 75.0 & 64.0 &  80.8 & 70.0 & 75.8 \\
        AE-CSA (\citeyear{luo2019unsupervised}) & 90.3 & 92.6 & 91.4 & 92.6 & 75.6 & 77.3 & 91.4 & 77.9 & 77.0  &  85.6 & 86.0 & 85.8 \\
        CAt (\citeyear{tulkens-van-cranenburgh-2020-embarrassingly}) & 91.8 & 92.4 & 92.1  & 82.4 & 75.6 & 78.8  &76.6 &80.1 &76.6 &   86.5 & 86.4 & 86.4 \\\hline
        ABAE + HRSMap & 93.0 & 88.8 & 90.9 & 85.8 & 75.3 & 80.2 & 67.4 & 89.6 & 76.9 &   87.0 & 85.8 & 86.0 \\
        SSCL & 91.7 & 94.6 & 93.1 & 88.4 & 75.9 & 81.7 & 79.1 & 86.1 & 82.4 &   88.8 & 88.7 & 88.6 \\
        SSCLS-BERT & 89.6 & 97.3 & 93.3 & 95.5 & 71.9 & 82.0  & 84.0 & 87.6 & 85.8  &  90.0 & 89.7 & 89.4 \\
        SSCLS-DistilBERT & 91.3 & 96.6 & \bf 93.9 & 92.4 & 75.9 & \bf 83.3 & 84.4 & 88.0 & \bf 86.2 &  90.4 & 90.3 & \bf 90.1 \\
        \bottomrule
    \end{tabular}
    \caption{Aspect-level precision (\textbf{P}), recall (\textbf{R}), and F-scores (\textbf{F}) on the Citysearch testing set.
    For overall, we calculate weighted macro averages across all aspects.}
    \label{tab:citysearch}
    \vspace{-3mm}
\end{table*}

\subsection{Comparison Methods}

We compare our methods against five baselines on the Citysearch dataset.
\textbf{SERBM} \cite{wang-etal-2015-sentiment} is a sentiment-aspect extraction restricted Boltzmann machine, which jointly extracts review aspects and sentiment polarities in an unsupervised manner.
\textbf{W2VLDA} \cite{GARCIAPABLOS2018127} is a topic modeling based approach, which combines word embeddings \cite{mikolov2013distributed} with Latent Dirichlet Allocation \cite{blei2003latent}.
It automatically pairs discovered topics with pre-defined aspect names based on user provided seed-words for different aspects.
\textbf{ABAE} \cite{he-etal-2017-unsupervised} is an autoencoder that aims at learning highly coherent aspects by exploiting the distribution of word co-occurrences using neural word embeddings, and an attention mechanism that can put emphasis on aspect-related keywords in segments during training.
\textbf{AE-CSA} \cite{luo2019unsupervised} improves ABAE by leveraging sememes to enhance lexical semantics, where sememes are obtained via WordNet \cite{miller1995wordnet}.
\textbf{CAt} \cite{tulkens-van-cranenburgh-2020-embarrassingly} is a simple heuristic model that consists of a contrastive attention mechanism based on Radial Basis Function kernels and an automated aspect assignment method.

For Amazon reviews, we compare our methods with several weakly supervised baselines, which explicitly leverage seed words extracted from human annotated development sets \cite{karamanolakis-etal-2019-leveraging} as supervision for aspect detection.
\textbf{ABAE$_{init}$} \cite{angelidis-lapata-2018-summarizing} replaces each aspect embedding vector in ABAE with the corresponding centroid of seed word embeddings, and fixes aspect embedding vectors during training.
\textbf{MATE} \cite{angelidis-lapata-2018-summarizing} uses the weighted average of seed word embeddings to initialize aspect embeddings.
\textbf{MATE-MT} extends MATE by introducing an additional multi-task training objective.
\textbf{TS-*}  \cite{karamanolakis-etal-2019-leveraging} is a weakly supervised student-teacher co-training framework, where \textbf{TS-Teacher} is a bag-of-words classifier (teacher) based on seed words.
\textbf{TS-Stu-W2V} and \textbf{TS-Stu-BERT}
are student networks that use word2vec embeddings and the BERT model to encode text segments, respectively.

\subsection{Implementation Details}

We implemented all deep learning models using PyTorch \cite{paszke2017automatic}.
For each dataset, the best parameters and hyperparameters are selected based on the development set.

For our SSCL model, word embeddings are pre-loaded with 128-dimensional word vectors trained by skip-gram model \cite{mikolov2013distributed} with negative sampling and fixed during training.
For each dataset, we use gensim\footnote{\url{https://radimrehurek.com/gensim/}} to train word embeddings from scratch and set both window and negative sample size to 5. 
The aspect embedding matrix is initialized with the centroids of clusters by running k-means on word embeddings.
We set the number of aspects to 30 for all datasets because the model can achieve competitive performance while it will still be relatively easier to map model-inferred aspects to gold-standard aspects.
The smooth factor $\lambda$ is tuned in $\{0.5, 1.0, 2.0, 3.0, 4.0, 5.0\}$ and set to $0.5$ for all datasets.
The temperature $\mu$ is set to 1.
For SSCLS, we have experimented with two pretrained encoders, i.e., BERT \cite{devlin2019bert} and DistilBERT \cite{sanh2019distilbert}.
We tune smooth factor $\lambda$ in $\{0.5, 1.0\}$, $\chi_\text{G}$ in $\{0.7, 0.8, 1.0, 1.2\}$, and $\chi_\text{NG}$ in $\{1.4, 1.6, 1.8\}$.
We set $\chi_\text{G}<\chi_\text{NG}$ to alleviate the label imbalance problem, since the majority of sentences in the corpus are labeled as \textit{General}.

For both SSCL and SSCLS, model parameters are optimized using the Adam optimizer \cite{kingma2014adam} with $\beta_1=0.9, \beta_2=0.999$, and $\epsilon=10^{-8}$.
Batch size is set to 50.
For learning rates, we adopt a warmup schedule strategy proposed in \cite{vaswani2017attention}, and set warmup step to 2000 and model size to $10^5$.
Gradient clipping with a threshold of 2 has also been applied to prevent gradient explosion.
Our codes are available at \url{https://github.com/tshi04/AspDecSSCL}.

\begin{table}[!t]
    \centering
    \resizebox{\linewidth}{!}{
    \begin{tabular}{l|m{18em}}
    \toprule
    \bf Aspects & \bf Representative Keywords \\
    \hline
    Apps/Interface 
    & apps app netflix browser hulu youtube \\\hline
    \multirow{2}{*}{Connectivity}
    & channel antenna broadcast signal station \\\cline{2-2}
    & optical composite hdmi input component \\\hline
    \multirow{2}{*}{Customer Serv.}
    & service process company contact support \\\cline{2-2}
    & call email contacted rep phone repair \\\hline
    Ease of Use 
    & button remote keyboard control use qwerty \\\hline
    \multirow{2}{*}{Image}
    & setting brightness mode contrast color \\\cline{2-2}
    & motion scene blur action movement effect \\\hline
    Price 
    & dollar cost buck 00 pay tax \\\hline
    Size/Look 
    & 32 42 37 46 55 40 \\\hline
    Sound 
    & speaker bass surround volume sound stereo \\\hline
    \multirow{7}{*}{General}
    & forum read reading review cnet posted \\\cline{2-2}
    & recommend research buy purchase decision \\\cline{2-2}
    & plastic glass screw piece metal base \\\cline{2-2}
    & foot wall mount stand angle cabinet \\\cline{2-2}
    & football watch movie kid night game \\\cline{2-2}
    & pc xbox dvd ps3 file game \\\cline{2-2}
    & series model projection plasma led sony \\
    \bottomrule
    \end{tabular}}
    \caption{Left: Gold-standard aspects for TVs reviews. Right: Model-inferred aspects presented by representative words.}
    \label{tab:aspect-words}
    \vspace{-6mm}
\end{table}

\subsection{Performance on Amazon Product Reviews}

Following previous works \cite{angelidis-lapata-2018-summarizing,karamanolakis-etal-2019-leveraging}, we use micro-averaged F1 score as our evaluation metric to measure the aspect detection performance among different models on Amazon product reviews.
All results are shown in Table~\ref{tab:perform_amazon}, where we use \textbf{bold} font to highlight the best performance values.
The results of the compared models are obtained from the corresponding published papers. From this table, we can observe that weakly supervised ABAE$_{init}$, MATE and MATE-MT perform significantly better than unsupervised ABAE since they leverage aspect representative words extracted from human-annotated datasets and thus leads to more accurate aspect predictions. TS-Teacher outperforms MATE and MATE-MT on most of the datasets, which further demonstrates that these words are highly correlated with gold-standard aspects.
The better performance of both TS-Stu-W2V and TS-Stu-BERT over TS-Teacher demonstrates the effectiveness of their teacher-student co-training framework.

In our experiments, we conjecture that low-resolution many-to-one aspect mapping may be one of the reasons for the low performance of traditional ABAE. Therefore, we have reimplemented ABAE and combined it with HRSMap.
The new model (i.e., ABAE + HRSMap) obtains significantly better results compared to the traditional ABAE on all datasets (performance improvement of $51.7\%$), showing HRSMap is effective in mapping model-inferred aspects to gold-standard aspects.
Compared to the TS-* baseline methods, our SSCL achieves better results on Boots, KBs, and TVs, and competitive results on Bags, B/T, and VCs.
On average, it outperforms TS-Teacher, TS-Stu-W2V, and TS-Stu-BERT by $16.9\%$, $3.9\%$, and $1.3\%$, respectively.
SSCLS-BERT and SSCLS-DistilBERT further boost the performance of SSCL by $5.4\%$ and $4.4\%$, respectively, thus demonstrating that knowledge distillation is effective in improving the quality of aspect prediction.

\subsection{Performance on Restaurant Reviews}

We have conducted more detailed comparisons on the Citysearch dataset, which has been widely used to benchmark aspect detection models. Following previous work \cite{tulkens-van-cranenburgh-2020-embarrassingly}, we use weighted macro averaged precision, recall and F1 score as metrics to evaluate the overall performance.
We also evaluate performance of different models for three major individual aspects by measuring aspect-level precision, recall, and F1 scores.
Experimental results are presented in Table~\ref{tab:citysearch}.
Results of compared models are obtained from the corresponding published papers.

From Table~\ref{tab:citysearch}, we also observe that ABAE + HRSMap performs significantly better than traditional ABAE.
Our SSCL outperforms all baselines in terms of weighted macro averaged F1 score.
SSCLS-BERT and SSCLS-DistilBERT further improve the performance of SSCL, and SSCLS-DistilBERT achieves the best results.
From aspect-level results, we can observe that, for each individual aspect, our SSCL, SSCLS-BERT and SSCLS-DistilBERT performs consistently better than compared baseline methods in terms of F1 score.
SSCLS-DistilBERT gets the best F1 scores across all three aspects.
This experiment demonstrates the strength of the contrastive learning framework, HRSMap, and knowledge distillation, which are able to capture high-quality aspects, effectively map model-inferred aspects to gold-standard aspects, and accurately predict aspect labels for the given segments.

\subsection{Aspect Interpretation}

\begin{table}[!t]
    \centering
    \resizebox{\linewidth}{!}{
    \begin{tabular}{l|m{18em}}
    \toprule
    \bf Aspects & \bf Representative Keywords \\
    \hline
    Battery & charge recharge life standby battery drain \\\hline
    Comfort & uncomfortable hurt sore comfortable tight pressure \\\hline
    \multirow{2}{*}{Connectivity} & usb cable charger adapter port ac\\\cline{2-2}
    & paired htc galaxy android macbook connected \\\hline
    Durability & minute hour foot day min second\\\hline
    Ease of Use & button pause track control press forward\\\hline
    Look & red light blinking flashing color blink\\\hline
    Price & 00 buck spend paid dollar cost \\\hline
    \multirow{3}{*}{Sound} & bass high level low treble frequency\\\cline{2-2}
    & noisy wind environment noise truck background\\\hline
    \multirow{10}{*}{General} & rating flaw consider star design improvement \\\cline{2-2}
    & christmas gift son birthday 2013 new husband \\\cline{2-2}
    & warranty refund shipping contacted sent email\\\cline{2-2}
    & motorola model plantronics voyager backbeat jabra \\\cline{2-2}
    & gym walk house treadmill yard kitchen \\\cline{2-2}
    & player video listen streaming movie pandora \\\cline{2-2}
    & read reading website manual web review \\\cline{2-2}
    & purchased bought buying ordered buy purchase \\
    \bottomrule
    \end{tabular}}
    \caption{Left: Gold-standard aspects for Bluetooth Headsets reviews. Right: Model inferred aspects presented by representative words.}
    \label{tab:kwds_bluetooth}
    \vspace{-6mm}
\end{table}

As SSCL achieves promising performance quantitatively on aspect detection compared to the baselines, we further show some qualitative results to interpret extracted concepts.
From Table~\ref{tab:aspect-words}, we notice that there is at least one model-inferred aspect corresponding to each of the gold-standard aspects, which indicates model-inferred aspects based on HRSMap have a good coverage.
We also find that model-inferred concepts, which are mapped to non-general gold-standard aspects, are fine-grained, and their representative words are meaningful and coherent.
For example, it is easy to map ``\textit{app, netflix, browser, hulu, youtube}'' to \textit{Apps/Interface}.
Compared to weakly supervised methods (such as MATE), SSCL is also able to discover new concepts.
For example, for aspects mapped to \textit{General}, we may label ``\textit{pc, xbox, dvd, ps3, file, game}'' as \textit{Connected Devices}, and ``\textit{plastic glass screw piece metal base}'' as \textit{Build Quality}.
Similarly, we observe that model-inferred aspects based on Bluetooth Headsets reviews also have sufficient coverage for gold-standard aspects (see Table~\ref{tab:kwds_bluetooth}).
We can easily map model inferred aspects to gold-standard ones since their keywords are meaningful and coherent.
For instance, it is obvious that ``\textit{red, light, blinking, flashing, color, blink}'' are related to \textit{Look} and ``\textit{charge, recharge, life, standby, battery, drain}'' are about \textit{Battery}.
For new aspect detection, ``\textit{motorola, model, plantronics, voyager, backbeatjabra}'' can be interpreted as \textit{Brand}.
``\textit{player, video, listen, streaming, movie, pandora}'' are about \textit{Usage}.

\subsection{Ablation Study and Parameter Sensitivity}

\begin{figure}[!t]
	\centering
	\begin{subfigure}[b]{0.48\linewidth}
		\includegraphics[width=\linewidth]{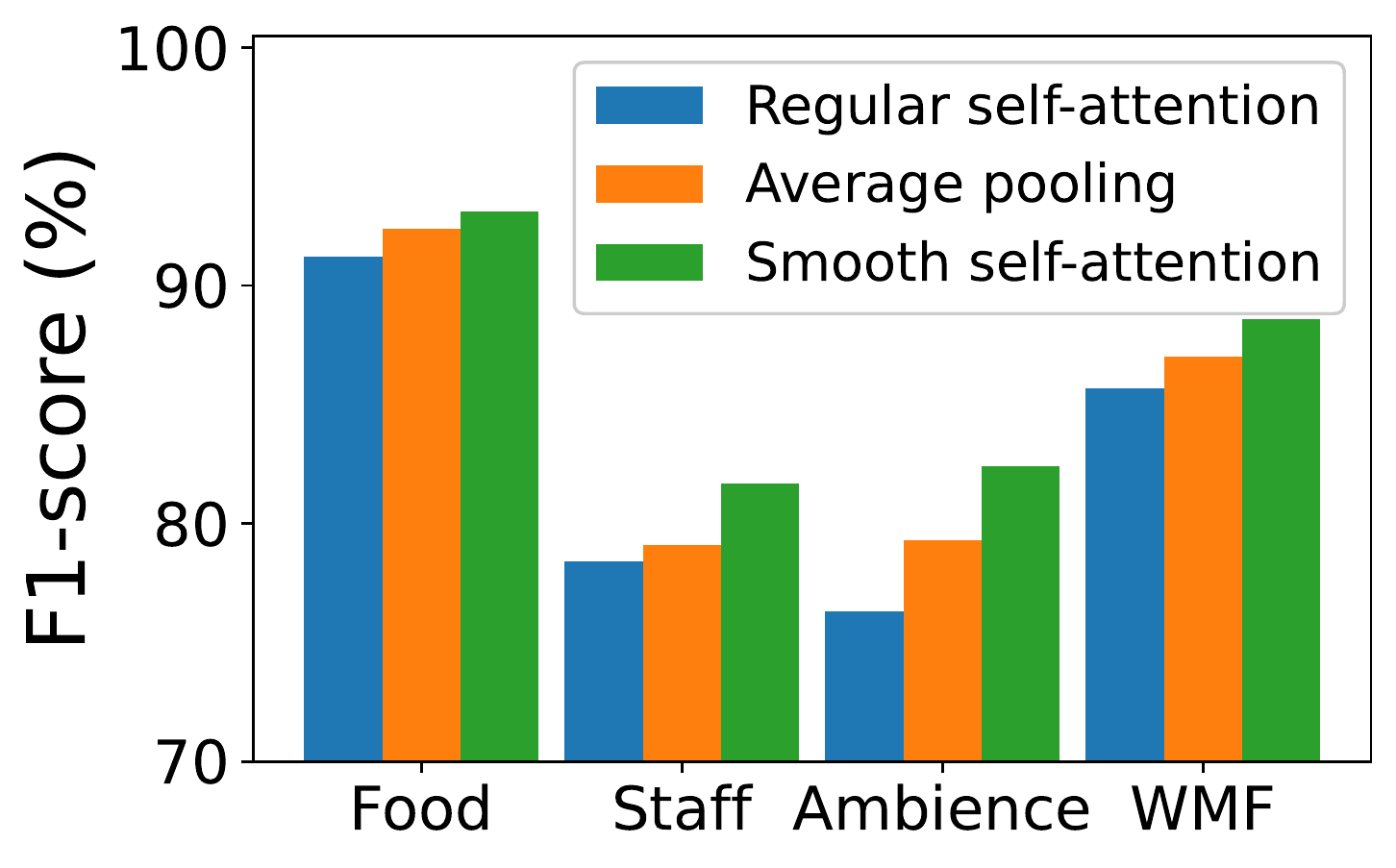}
		\caption{Smooth Self-Attention}
	\end{subfigure}
	\begin{subfigure}[b]{0.48\linewidth}
		\includegraphics[width=\linewidth]{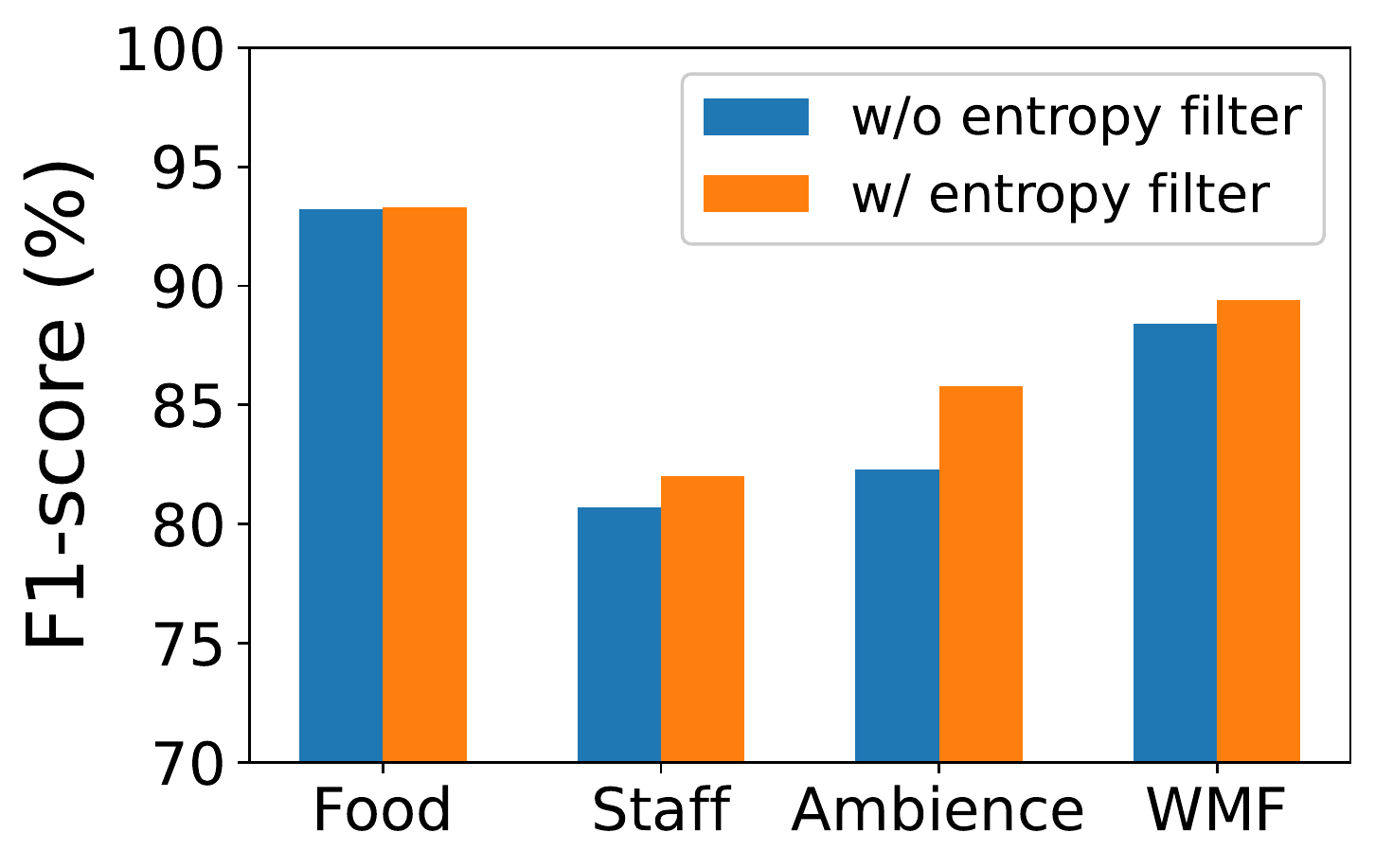}
		\caption{Entropy Filter}
	\end{subfigure}
	\caption{Ablation study on the Citysearch testing set. \textbf{WMF} represents weighted macro averaged F1-score.}
	\label{fig:ablation}
	\vspace{-3mm}
\end{figure}

\begin{figure}[!t]
	\centering
	\begin{subfigure}[b]{0.48\linewidth}
		\includegraphics[width=\linewidth]{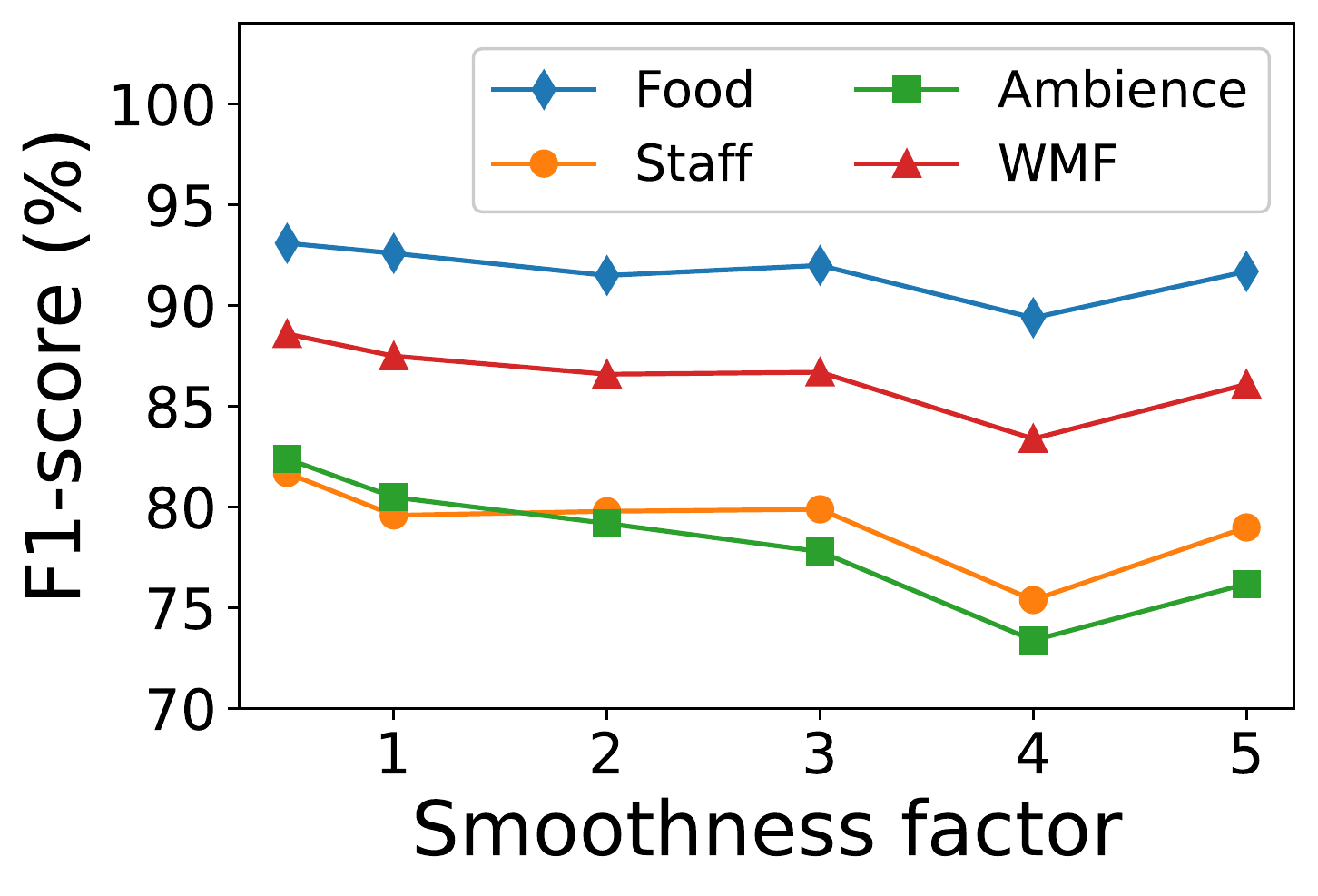}
		\caption{F1 vs. smoothness factor $\lambda$}
	\end{subfigure}
	\begin{subfigure}[b]{0.48\linewidth}
		\includegraphics[width=\linewidth]{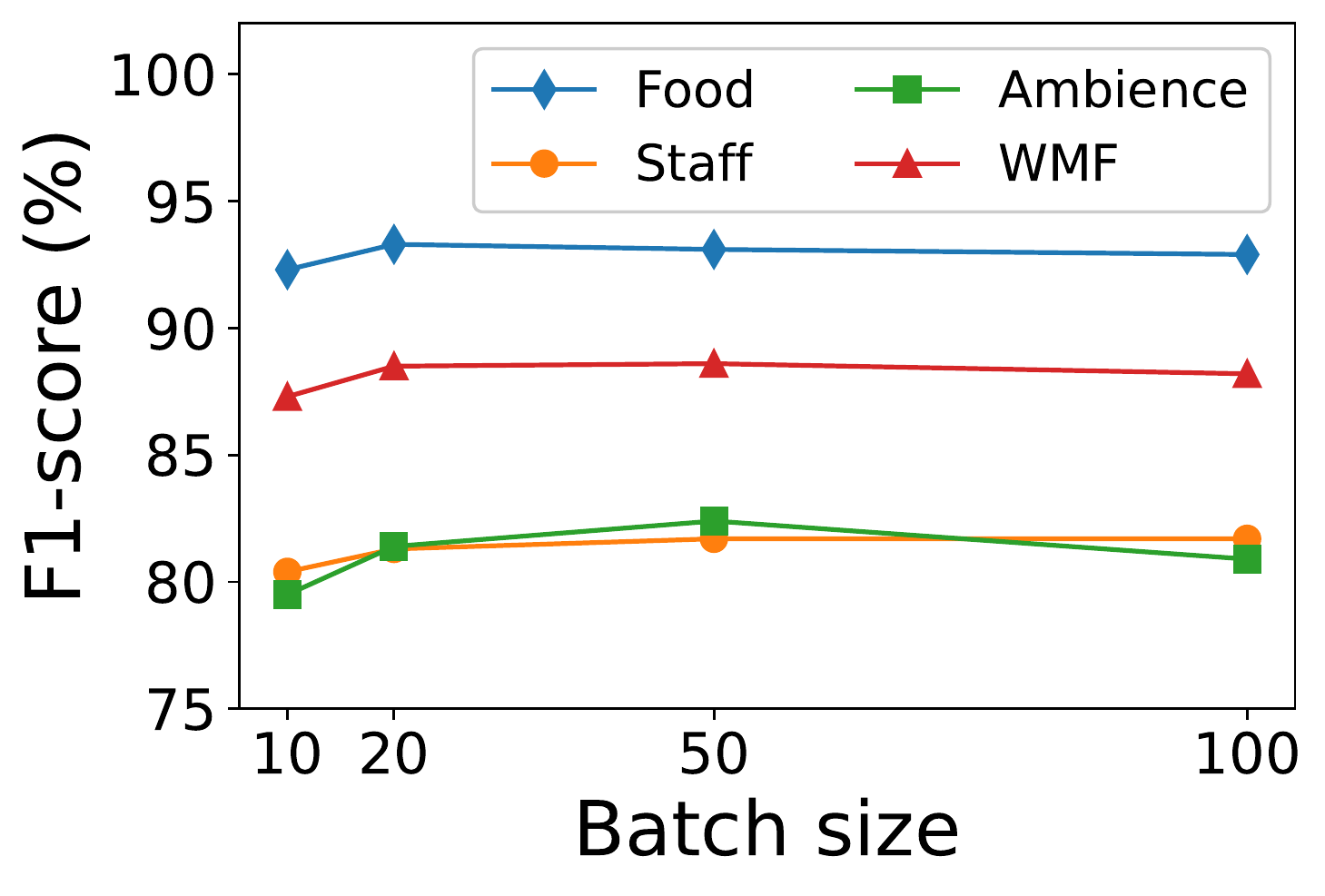}
		\caption{F1 vs. Batch Size}
	\end{subfigure}
	\caption{Parameter sensitivity analysis on Citysearch.}
	\label{fig:parameter}
	\vspace{-6mm}
\end{figure}

In addition to self-supervised contrastive learning framework and HRSMap, we also attribute the promising performance of our models to (i) Smooth self-attention mechanism, (ii) Entropy filters, and (iii) Appropriate batch size. Hence, we systematically conduct ablation studies and parameter sensitivity analysis to demonstrate the effectiveness of them, and provide the results in Fig.~\ref{fig:ablation} and Fig.~\ref{fig:parameter}.

First, we replace the smooth self-attention (SSA) layer with a regular self-attention (RSA) layer used in \cite{angelidis-lapata-2018-summarizing} and an average pooling (AP) layer.
The model with SSA performs better than the one with AP or RSA.
Next, we examine the entropy filter for SSCLS-BERT, and observe that adding it has a positive impact on the model performance.
Then, we study the effect of smoothness factor $\lambda$ in SSA and observe that our model achieves promising and stable results when $\lambda\leq 1$.
Finally, we investigate the effect of batch size.
F1 scores increase with batch size and become stable when batch size is greater than 20.
However, very large batch size increases the computational complexity; see Algorithm~\ref{alg:sscl}.
Therefore, we set batch size to 50 for all our experiments.

\subsection{Case Study}

\begin{figure}[!t]
	\centering
	\includegraphics[width=0.65\linewidth]{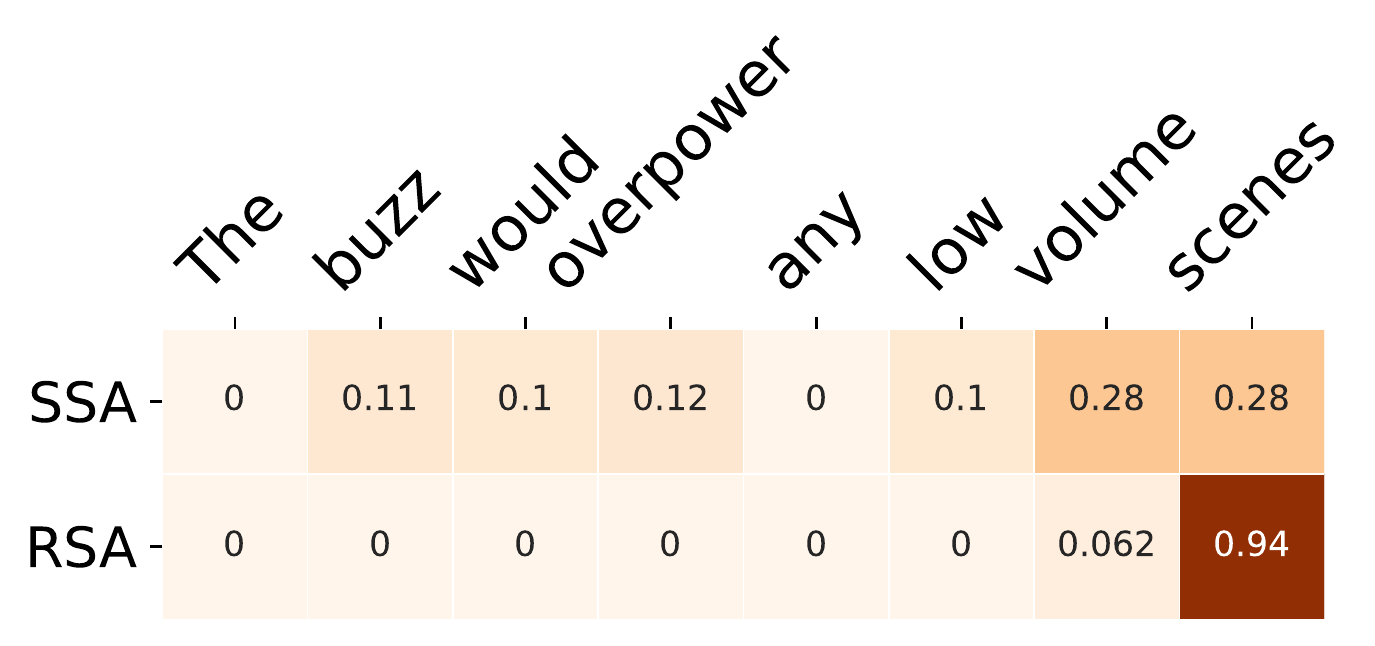}
	\includegraphics[width=0.65\linewidth]{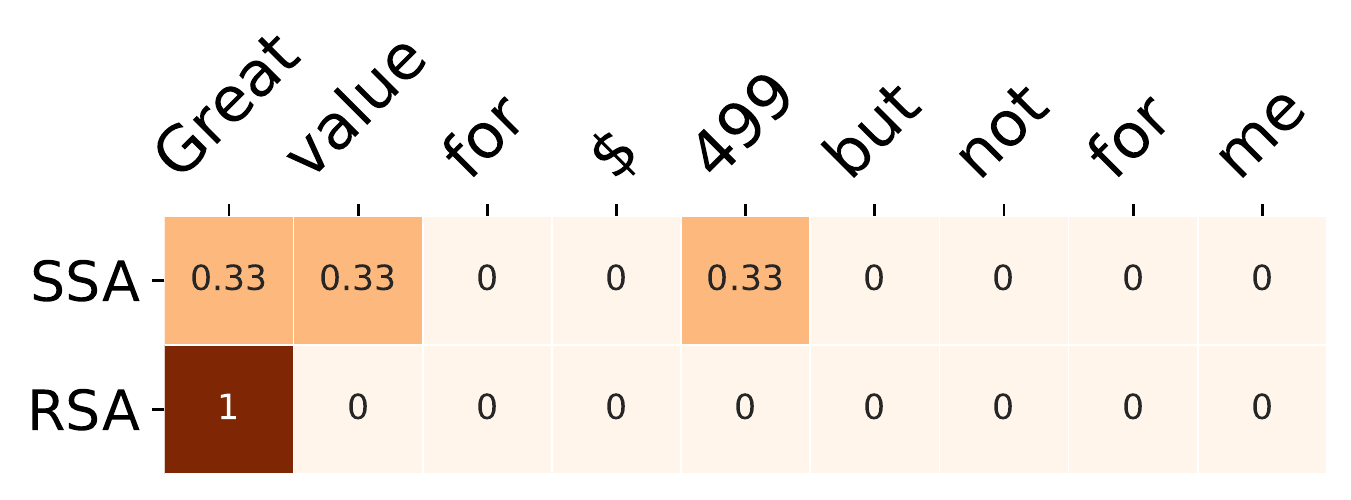}
	\caption{
	Visualization of attention weights. SSA and RSA represent smooth and regular self-attention, respectively.}
	\label{fig:vis_attn}
	\vspace{-3mm}
\end{figure}

Fig.~\ref{fig:vis_attn} compares heat-maps of attention weights obtained from SSA and RSA on two segments from the Amazon TVs testing set.
In each example, RSA attempts to use a single word to represent the entire segment. However, the word may be either a representative word for another aspect (e.g., ``\textit{scene}'' for \textit{Image} in Table~\ref{tab:aspect-words}) or a word with no aspect tendency (e.g., ``\textit{great}'' is not assigned to any aspect).
In contrast, SSA captures phrases and multiple words, e.g., ``\textit{volume scenes}'' and ``\textit{great value, 499}''.
Based on the results in Fig.~\ref{fig:ablation} and 
Fig.~\ref{fig:vis_attn}, we argue SSA is more robust and intuitively meaningful than RSA for aspect detection.

%% file: conclusion.tex
\vspace{-1mm}
\section{Conclusion}
\label{sec:conclusion}

In this paper, we propose a self-supervised contrastive learning framework for aspect detection.
Our model is equipped with two attention modules, which allows us to represent every segment with word embeddings and aspect embeddings, so that we can map aspect embeddings to the word embedding space through a contrastive learning mechanism.
In the attention module over word embeddings, we introduce a SSA mechanism.
Thus, our model can learn robust representations, since SSA encourages the model to capture phrases and multiple keywords in the segments.
In addition, we propose a HRSMap method for aspect mapping, which dramatically increases the accuracy of segment aspect predictions for both ABAE and our model.
Finally, we further improve the performance of aspect detection through knowledge distillation.
BERT-based student models can benefit from pretrained encoders and overcome the disadvantages of data preprocessing for the teacher model.
During training, we introduce entropy filters in the loss function to ensure student models focus on high confidence training samples.
Our models have shown better performance compared to several recent unsupervised and weakly-supervised models on several publicly available review datasets across different domains.
Aspect interpretation results show that extracted aspects are meaningful, have a good coverage, and can be easily mapped to gold-standard aspects.
Ablation studies and visualization of attention weights further demonstrate the effectiveness of SSA and entropy filters.

%% file: appendix.tex
\section{Supplementary Materials}

\setcounter{table}{0}
\renewcommand{\thetable}{A\arabic{table}}

\subsection{Datasets}

In this section, we provide more details about the datasets used in our experiments.

\subsubsection{Amazon Reviews}

We obtain Amazon product reviews from the OPOSUM dataset \cite{angelidis-lapata-2018-summarizing}, which has six subsets across different domains, including Laptop Cases, Bluetooth Headsets, Boots, Keyboards, Televisions, and Vacuums.
For each subset, reviews are segmented into elementary discourse units (EDUs) through a Rhetorical Structure Theory parser \cite{feng2014linear}.
Then, each segment in development and test sets is manually annotated with eight representative aspect labels as well as aspect \textit{General}.
We show the annotated aspect labels in Table~\ref{tab:anno_aspects}.
In our experiments, we use exactly the same segments and aspect labels as \cite{angelidis-lapata-2018-summarizing}.

\subsubsection{Restaurant Reviews}
For restaurant reviews, training and testing sets are from the Citysearch dataset \cite{he-etal-2017-unsupervised}, while the development set is a combination of restaurant subsets of SemEval 2014 and SemEval 2015 Aspect-Based Sentiment Analysis datasets \cite{pontiki-etal-2014-semeval, pontiki-etal-2015-semeval}.
Similar to previous work \cite{he-etal-2017-unsupervised}, sentences are treated as segments.
In the development and testing sets, we select sentences that only express one aspect, and disregard those with multiple and no aspect labels. 
We have also restricted ourselves to three labels (i.e., \textit{Food}, \textit{Service}, and \textit{Ambience}), to form a fair comparison with prior work \cite{he-etal-2017-unsupervised,tulkens-van-cranenburgh-2020-embarrassingly}.

In our experiments, we have also exploited the English restaurant review dataset from SemEval-2016 Aspect-based Sentiment Analysis task \cite{pontiki2016semeval} containing reviews for multiple domains and languages, which has been used in prior work \cite{karamanolakis-etal-2019-leveraging} for aspect detection.
However, we find that the dataset suffers from severe label-imbalance problem.
For example, there are only 3 and 13 out of 676 sentences labeled as \textit{drinks\#prices} and \textit{location\#general}, respectively.

\subsection{Aspect Mapping}

In this section, we provide more details of high-resolution selective mapping (HRSMap).
High-resolution refers to the fact that the number of model-inferred aspects \textbf{(MIAs)} should be at least 3 times more than the number of gold-standard aspects \textbf{(GSAs)}, so that model-inferred aspects have a better coverage. 
Selective mapping implies that noisy or meaningless aspects will not be mapped to gold-standard aspects.

In our experiments, we set the number of MIAs to 30, considering the balance between aspect coverage and human-effort to manually map them to gold-standard aspects.
Usually, it takes less than 15 minutes to assign 30 MIAs to GSAs.
First, we automatically generate keywords of MIAs and dump them into a text file, where the number of the most relevant keywords for each aspect is 10.
Second, we create several rules for aspect mapping: (i) If keywords of a MIA are clearly related to one specific GSA (not \textit{General}), we map this MIA to the GSA.
For example, we map ``\textit{apps, app, netflix, browser, hulu, youtube, stream}'' to \textit{Apps/Interface}.
(ii) If keywords are coherent but not related to any specific GSA, we map this MIA to \textit{General}.
For instance, we map ``\textit{football, watch, movie, kid, night, family}'' to \textit{General}.
(iii) If keywords are related to more than one GSA, we treat this MIA as a noisy aspect and it will not be mapped.
For example, ``\textit{excellent, amazing, good, great, outstanding, fantastic, impressed, superior}'' may be related to several different GSAs.
(iv) If keywords are not quite meaningful, their corresponding MIA will not be mapped.
For instance, ``\textit{ago, within, last 30, later, took, couple, per, every}'' is a meaningless MIA.
Third, we further verify the quality of aspect mapping using development sets.

We provide more qualitative results to demonstrate: 
(i) MIAs are meaningful and interpretable. 
(ii) MIAs based on HRSMap have good coverage. 
(iii) Our model is able to discover new aspects.
All results are summarized in Tables~\ref{tab:kwds_bags}, \ref{tab:kwds_boots}, \ref{tab:kwds_keyboards}, \ref{tab:kwds_vacuums}, and \ref{tab:kwds_restaurant}.

\begin{table}[!t]
    \centering
    \resizebox{\linewidth}{!}{
    \begin{tabular}{l|m{18em}}
    \toprule
    \bf Aspects & \bf Representitive Keywords \\\hline
    Compartments & zippered velcro flap main zipper front\\\hline
    \multirow{2}{*}{Customer Serv.} & service customer warranty shipping contacted email\\\cline{2-2}
    & shipping arrived return shipped sent amazon \\\hline
    Handles & shoulder strap chest comfortable weight waist \\\hline
    Looks & color blue pink purple green bright\\\hline
    Price & 50 cost spend paid dollar price\\\hline
    Protection & protect protection protects protecting protected safe\\\hline
    \multirow{2}{*}{Quality} & scratch dust drop damage scratched bump\\\cline{2-2}
    & material plastic fabric soft foam leather\\\hline
    \multirow{3}{*}{Size/Fit} & inch perfectly snug tight dell nicely\\\cline{2-2}
    & plenty lot amount enough ton extra\\\cline{2-2}
    & 17 15 13 14 11 16\\\hline
    \multirow{6}{*}{General} & purchased bought ordered buying buy owned \\\cline{2-2}
    & review read people mentioned reviewer reading \\\cline{2-2}
    & airport security tsa friendly checkpoint luggage \\\cline{2-2}
    & trip travel carry seat traveling school \\
    \bottomrule
    \end{tabular}}
    \caption{Left: GSAs for Laptop Cases reviews. Right: MIAs presented by representative words.}
    \label{tab:kwds_bags}
    \vspace{-3mm}
\end{table}


\begin{table}[!t]
    \centering
    \resizebox{\linewidth}{!}{
    \begin{tabular}{l|m{18em}}
    \toprule
    \bf Aspects & \bf Representative Keywords \\
    \hline
    Color & color darker brown dark grey gray\\\hline
    \multirow{2}{*}{Comfort} & calf leg ankle shaft top knee \\\cline{2-2}
    & hurt blister pain sore break rub \\\hline
    Durability & ago wore apart worn started last \\\hline
    Look & casual stylish cute compliment dressy sexy \\\hline
    \multirow{2}{*}{Materials} & slippery traction sole grip tread rubber \\\cline{2-2}
    & insole lining insert wool liner padding \\\hline
    Price & price paid pay spend cost money \\\hline
    \multirow{2}{*}{Size} & 16 13 14 knee circumference 15 \\\cline{2-2}
    & room large big wide tight bigger \\\hline
    Weather Resist. & snow dry water cold wet weather \\\hline
    \multirow{10}{*}{General} & box rubbed weird near cut make \\\cline{2-2}
    & brand owned miz marten mooz clark \\\cline{2-2}
    & walking walk floor office town walked \\\cline{2-2}
    & christmas store local gift daughter birthday \\\cline{2-2}
    & suggest recommend buy probably consider thinking \\\cline{2-2}
    & amazon best description future satisfied needle \\\cline{2-2}
    & reviewer review people others everyone someone \\\cline{2-2}
    & shipping service seller return delivery amazon \\
    \bottomrule
    \end{tabular}}
    \caption{Left: GSAs for Boots reviews. Right: MIAs presented by representative words.}
    \label{tab:kwds_boots}
    \vspace{-3mm}
\end{table}

\begin{table}[!t]
    \centering
    \resizebox{\linewidth}{!}{
    \begin{tabular}{l|m{18em}}
    \toprule
    \bf Aspects & \bf Representative Keywords \\
    \hline
    Build Qual. & plastic case stand cover bag angle \\\hline
    Connectivity & cable port receiver cord usb dongle \\\hline
    Extra Func. & volume pause mute medium music player \\\hline
    Feel Comfort & wrist hand pain easier typing finger \\\hline
    \multirow{2}{*}{Layout} & smaller size larger sized layout bigger \\\cline{2-2}
    & backspace shift delete fn arrow alt \\\hline
    \multirow{2}{*}{Looks} & black white see finish color wear lettering print show glossy \\\cline{2-2}
    & lighting light color bright lit dark \\\hline
    Noise & feedback tactile cherry sound loud noise \\\hline
    Price & price cost dollar buck pay money \\\hline
    \multirow{10}{*}{General} 
    & galaxy tablet pair ipad samsung android \\\cline{2-2}
    & web email text video movie document  \\\cline{2-2}
    & microsoft ibm natural purchased hp dell \\\cline{2-2}
    & amazon sent customer seller contacted service \\\cline{2-2}
    & driver software window install download \\\cline{2-2}
    & recommend buy highly purchase gaming  buying \\\cline{2-2}
    & month week stopped ago year started \\\cline{2-2}
    & room couch tv living pc desk \\\cline{2-2}
    & star negative flaw complain complaint review \\
    \bottomrule
    \end{tabular}}
    \caption{Left: GSAs for Keyboards reviews. Right: MIAs presented by representative words.}
    \label{tab:kwds_keyboards}
\end{table}

\begin{table}[!t]
    \centering
    \resizebox{\linewidth}{!}{
    \begin{tabular}{l|m{18em}}
    \toprule
    \bf Aspects & \bf Representative Keywords \\
    \hline
    \multirow{2}{*}{Accessories}
    & extension powered turbo tool attachment accessory \\\cline{2-2}
    & container cup bin bag canister tank \\\hline
    Build Quality 
    & plastic screw clip tube tape hose \\\hline
    Customer Serv.
    & repair warranty send service called contacted \\\hline
    Ease of Use 
    & height switch button setting adjust turn \\\hline
    Noise 
    & difference quality noise design sound flaw \\\hline
    Price 
    & 00 cost dollar buck paid shipping \\\hline
    Suction Power
    & crumb food litter hair sand fur \\\hline
    Weight 
    & easier difficult heavy awkward cumbersome lug \\\hline
    \multirow{9}{*}{General}
    & recommend thinking suggest money regret thought \\\cline{2-2}
    & read mentioned reading negative agree complained \\\cline{2-2}
    & purchased bought buying ordered buy purchasing \\\cline{2-2}
    & died lasted broke stopped within last \\\cline{2-2}
    & eureka kenmore electrolux hoover model upright \\\cline{2-2}
    & corner table bed ceiling chair furniture \\
    \bottomrule
    \end{tabular}}
    \caption{Left: GSAs for Vacuums reviews. Right: MIAs presented by representative words.}
    \label{tab:kwds_vacuums}
    \vspace{-2mm}
\end{table}

\begin{table}[!t]
    \centering
    \resizebox{\linewidth}{!}{
    \begin{tabular}{l|m{18em}}
    \toprule
    \bf Aspects & \bf Representative Keywords \\
    \hline
    \multirow{4}{*}{Ambience}
    & room wall ceiling wood floor window \\\cline{2-2}
    & music dj bar fun crowd band \\\cline{2-2}
    & atmosphere romantic cozy feel decor intimate \\\cline{2-2}
    & wall ceiling wood high black lit \\\hline
    \multirow{7}{*}{Food}
    & steak medium cooked fry dry tender \\\cline{2-2}
    & pork chicken potato goat rib roast \\\cline{2-2}
    & tuna shrimp pork lamb salmon duck \\\cline{2-2}
    & chocolate coffee cake cream tea dessert \\\cline{2-2}
    & large small big three four huge \\\cline{2-2}
    & tomato sauce cheese onion oil crust \\\cline{2-2}
    & american menu variety japanese italian cuisine \\\hline
    \multirow{3}{*}{Staff}
    & staff waiter server waitress waitstaff manager \\\cline{2-2}
    & friendly attentive helpful prompt knowledgeable courteous \\\hline
    \multirow{8}{*}{General}
    & per tip bill 20 fixe dollar \\\cline{2-2}
    & sunday night saturday friday weekend evening \\\cline{2-2}
    & ago birthday anniversary recently last celebrate \\\cline{2-2}
    & overpriced worth average quality bit pretty \\\cline{2-2}
    & street west east park manhattan village \\\cline{2-2}
    & minute year month min hour week \\\cline{2-2}
    & review say heard believe read reading \\
    \bottomrule
    \end{tabular}}
    \caption{Left: GSAs for Restaurant reviews. Right: MIAs presented by representative words.}
    \label{tab:kwds_restaurant}
\end{table}

\subsection{Ablation Study and Parameter Sensitivity}

In this section, we provide more results for ablation study and parameter sensitivity.
Tables~\ref{tab:ablation_amazon} and \ref{tab:ablation_restaurant} show models with SSA achieve better performance than those with RSA and AVGP.
Tables~\ref{tab:lambda_amazon} and \ref{tab:lambda_restaurant} show effects of the smoothness factor on the performance of our SSCL model.
We find that our model achieves promising and stable results when $\lambda\leq 1.0$ and $\lambda$ is fixed to 0.5 for all datasets.
From Table~\ref{tab:batch_size_amazon} and \ref{tab:batch_size_restaurant}, we can see that
F1 scores increase with batch size and become stable when batch size is greater than 20.
According to Algorithm~\ref{alg:sscl} line 7-8, we calculate similarities for $X^2$ times at each training step, where $X$ is the batch size.
Since large batch size requires extra computations, we set batch size to 50 for all our experiments as a trade-off between performance and computational complexity.

\begin{table}[!t]
    \centering
    \resizebox{\linewidth}{!}{
    \begin{tabular}{lccccccc}
    \toprule
    \bf Smooth & \bf Bags & \bf B/T & \bf Boots & \bf KBs & \bf TVs & \bf VCs & \bf AVG \\\hline
    SSA & 61.0 & 65.2 & 57.3 & 60.6 & 64.6 & 57.2 & 61.0 \\
    RSA & 55.9 & 62.3 & 52.9 & 59.5 & 59.5 & 53.9 & 57.3\\
    AVGP & 61.6 & 65.5 & 52.7 & 60.5 & 64.0 & 56.0 & 60.1 \\
    \bottomrule
    \end{tabular}}
    \caption{Effects of SSA on micro-averaged F1 scores for Amazon review datasets. SSA, RSA, AVGP represent smooth self-attention, regular self-attention and average-pooling, respectively.}
    \label{tab:ablation_amazon}
\end{table}

\begin{table}[!t]
    \centering
    \begin{tabular}{lcccc}
    \toprule
    \bf Smooth & \bf Food & \bf Staff & \bf Ambience & \bf WMF \\\hline
    SSA & 93.1 & 81.7 & 82.4 & 88.6 \\
    RSA & 91.2 & 78.4 & 76.3 & 85.7 \\
    AVGP & 92.4 & 79.1 & 79.3 & 87.0 \\
    \bottomrule
    \end{tabular}
    \caption{Effects of SSA on aspect-level F1 scores and weighted macro-averaged F1 scores for the Citysearch dataset. WMF represents weighted macro averaged F1-score.}
    \label{tab:ablation_restaurant}
\end{table}

\newpage

\begin{table}[!t]
    \centering
    \resizebox{\linewidth}{!}{
    \begin{tabular}{lccccccc}
    \toprule
    $\boldsymbol \lambda$ & \bf Bags & \bf B/T & \bf Boots & \bf KBs & \bf TVs & \bf VCs & \bf AVG \\\hline
    0.5 & 61.0 & 65.2 & 57.3 & 60.6 & 64.6 & 57.2 & 61.0 \\
    1.0 & 61.6 & 65.1 & 58.3 & 61.8 & 66.4 & 55.6 & 61.5 \\
    2.0 & 60.7 & 63.9 & 57.3 & 59.8 & 67.0 & 55.0 & 60.6 \\
    3.0 & 61.8 & 64.6 & 57.6 & 59.9 & 63.0 & 55.3 & 60.4 \\
    4.0 & 58.2 & 64.2 & 54.0 & 59.9 & 64.3 & 56.1 & 59.4 \\
    5.0 & 57.4 & 63.0 & 54.2 & 59.3 & 66.4 & 54.9 & 59.2 \\
    \bottomrule
    \end{tabular}}
    \caption{Effects of smoothness factor $\lambda$ on micro-averaged F1 scores for Amazon review datasets.}
    \label{tab:lambda_amazon}
\end{table}

\begin{table}[!t]
    \centering
    \begin{tabular}{lcccc}
    \toprule
    $\boldsymbol \lambda$ & \bf Food & \bf Staff & \bf Ambience & \bf WMF \\\hline
    0.5 & 93.1 & 81.7 & 82.4 & 88.6 \\
    1.0 & 92.6 & 79.6 & 80.5 & 87.5 \\
    2.0 & 91.5 & 79.8 & 79.2 & 86.6 \\
    3.0 & 92.0 & 79.9 & 77.8 & 86.7 \\
    4.0 & 89.4 & 75.4 & 73.4 & 83.4 \\
    5.0 & 91.7 & 79.0 & 76.2 & 86.1 \\
    \bottomrule
    \end{tabular}
    \caption{Effects of smoothness factor $\lambda$ on aspect-level F1 scores and weighted macro-averaged F1 scores for the Citysearch dataset.}
    \label{tab:lambda_restaurant}
\end{table}



\begin{table}[htp]
    \centering
    \resizebox{\linewidth}{!}{
    \begin{tabular}{lccccccc}
    \toprule
    \bf Bsize & \bf Bags & \bf B/T & \bf Boots & \bf KBs & \bf TVs & \bf VCs & \bf AVG \\\hline
    20 & 60.2 & 66.9 & 56.0 & 60.4 & 66.7 & 56.3 & 61.1 \\
    50 & 61.0 & 65.2 & 57.3 & 60.6 & 64.6 & 57.2 & 61.0 \\
    100 & 61.8 & 66.0 & 55.8 & 61.4 & 63.4 & 57.4 & 61.0 \\
    200 & 59.4 & 64.6 & 56.3 & 60.8 & 64.6 & 56.6 & 60.4 \\
    \bottomrule
    \end{tabular}}
    \caption{Effects of batch size on micro-averaged F1 scores for Amazon review datasets.}
    \label{tab:batch_size_amazon}
\end{table}

\begin{table}[htp]
    \centering
    \begin{tabular}{lcccc}
    \toprule
    \bf \bf Bsize & \bf Food & \bf Staff & \bf Ambience & \bf WMF \\\hline
    10 & 92.3 & 80.4 & 79.5 & 87.3 \\
    20 & 93.3 & 81.3 & 81.4 & 88.5 \\
    50 & 93.1 & 81.7 & 82.4 & 88.6 \\
    100 & 92.9 & 81.7 & 80.9 & 88.2 \\
    200 & 93.0 & 82.6 & 82.9 & 88.9 \\
    \bottomrule
    \end{tabular}
    \caption{Effects of batch size on aspect-level F1 scores and weighted macro-averaged F1 scores for the Citysearch dataset.}
    \label{tab:batch_size_restaurant}
\end{table}




